\documentclass[conference]{IEEEtran}
\IEEEoverridecommandlockouts
\usepackage{cite}
\usepackage{amsmath,amssymb,amsfonts}
\usepackage{algorithmic}
\usepackage{graphicx}
\usepackage{textcomp}
\usepackage{xcolor}

\usepackage{CJKutf8}
\usepackage{makecell}
\usepackage{subcaption}
\usepackage{xspace}
\usepackage{float}

\usepackage{tipa}

\usepackage{multirow}
\usepackage{multicol}
\usepackage{enumitem}
\usepackage{url}
\usepackage{colortbl}
\usepackage{tcolorbox}
\usepackage{booktabs}
\usepackage{balance}

\usepackage[T1]{fontenc}

\def\BibTeX{{\rm B\kern-.05em{\sc i\kern-.025em b}\kern-.08em
    T\kern-.1667em\lower.7ex\hbox{E}\kern-.125emX}}
\begin{document}

\title{MTTM: Metamorphic Testing for Textual  Content Moderation Software}

\author{
\IEEEauthorblockN{Wenxuan Wang\IEEEauthorrefmark{1}, Jen-tse Huang\IEEEauthorrefmark{1}, Weibin Wu\IEEEauthorrefmark{2}, Jianping Zhang\IEEEauthorrefmark{1}, Yizhan Huang\IEEEauthorrefmark{1}, Shuqing Li\IEEEauthorrefmark{1},\\Pinjia He\IEEEauthorrefmark{3}\thanks{Pinjia He is the corresponding author.}, and Michael R. Lyu\IEEEauthorrefmark{1}}
\IEEEauthorblockA{
    \IEEEauthorrefmark{1}
    Department of Computer Science and Engineering, The Chinese University of Hong Kong, Hong Kong, China\\
    \IEEEauthorrefmark{2}
    School of Software Engineering, Sun Yat-sen University, Zhuhai, China\\
    \IEEEauthorrefmark{3}
    School of Data Science, The Chinese University of Hong Kong, Shenzhen, Shenzhen, China\\
\{wxwang, jthuang, jpzhang, yzhuang22, sqli21, lyu\}@cse.cuhk.edu.hk\\
wuwb36@mail.sysu.edu.cn, hepinjia@cuhk.edu.cn}
}

\maketitle

\newcommand{\methodname}{MTTM\xspace}

\begin{abstract}
The exponential growth of social media platforms such as Twitter and Facebook has revolutionized textual communication and textual content publication in human society.
However, they have been increasingly exploited to propagate toxic content, such as hate speech, malicious advertisement, and pornography, which can lead to highly negative impacts (e.g., harmful effects on teen mental health).
Researchers and practitioners have been enthusiastically developing and extensively deploying textual content moderation software to address this problem.
However, we find that malicious users can evade moderation by changing only a few words in the toxic content. 
Moreover, modern content moderation software's performance against malicious inputs remains underexplored.
To this end, we propose \textit{\methodname}, a \underline{M}etamorphic \underline{T}esting framework for \underline{T}extual content \underline{M}oderation software.
Specifically, we conduct a pilot study on $2,000$ text messages collected from real users and summarize eleven metamorphic relations across three perturbation levels: character, word, and sentence.
\methodname employs these metamorphic relations on toxic textual contents to generate test cases, which are still toxic yet likely to evade moderation.
In our evaluation, we employ \methodname to test three commercial textual content moderation software and two state-of-the-art moderation algorithms against three kinds of toxic content.
The results show that \methodname achieves up to $83.9\%$, $51\%$, and $82.5\%$ error finding rates (EFR) when testing commercial moderation software provided by Google, Baidu, and Huawei, respectively, and it obtains up to $91.2\%$ EFR when testing the state-of-the-art algorithms from the academy.
In addition, we leverage the test cases generated by \methodname to retrain the model we explored, which largely improves model robustness ($0\%\sim5.9\%$ EFR) while maintaining the accuracy on the original test set.
A demo can be found in this link\footnote{http://ariselab.cse.cuhk.edu.hk/projects.html}.
\end{abstract}


\section{Introduction}
\label{sec-introduction}
In the recent decade, social media platforms and community forums have been developing rapidly, which tremendously facilitates modern textual communication and content publication worldwide.
For example, the number of tweets posted on Twitter has grown from $50$ million per day in $2010$ to $500$ million per day in $2020$~\cite{tweets2020}. 
However, they inevitably exacerbate the propagation of toxic content due to the anonymity of the web.
Textual toxic contents typically refer to three major kinds of texts: (1) \textit{abusive language and hate speech}, which are abusive texts targeting specific individuals, such as politicians, celebrities, religions, nations, and the LGBTIQA+ \cite{Badjatiya2017DeepLF}; (2) \textit{malicious advertisement}, which are online advertisements with illegal purposes, such as phishing and scam links, malware download, and illegal information dissemination~\cite{Li2012KnowingYE}; and (3) \textit{pornography}, which is often sexually explicit, associative, and aroused~\cite{Rowley2006LargeSI}.
These toxic contents can lead to highly negative impacts.
Specifically, Munro~\cite{children2011} studied the ill effects of online \textit{hate speech} on children and found that children may develop depression, anxiety, and other mental health problems.
\textit{Malicious advertisements} remain a notorious global burden, accounting for up to $85\%$ of daily message traffic \cite{spam2022}.
\textit{Pornography} can cause significant undesirable effects on the physical and psychological health of children \cite{Yu2016InternetMI}.
Moreover, these toxic contents can even increase the number of criminal cases to a certain extent~\cite{Chen2020AutomaticDO}.
All these studies reflect that toxic content can largely threaten social harmony; thus, content moderation software, which detects and blocks toxic content, has attracted massive interest from academia and industry.

Typical content moderation software first detects toxic content and then blocks it or warns the users before showing it.
As the core of content moderation, toxic content detection has been widely formulated as a classification task, and it has been tackled by various deep learning models, such as convolutional neuron networks, Long-Short-Term-Memory (LSTM) models, and Transformer models \cite{Mishra2019TacklingOA, Schmidt2017ASO, Wu2018TwitterSD}.
Recently, the development of pre-trained language models (e.g., BERT~\cite{Devlin2019BERTPO} and RoBERTa~\cite{Liu2019RoBERTaAR}) has significantly improved the held-out accuracy of toxic content detection.
Because of the recent progress in this field, industrial companies have also extensively deployed commercial-level content moderation software on their products, such as Google~\cite{google2021}, Facebook~\cite{facebook2020}, Twitter~\cite{twitter2020}, and Baidu~\cite{notrobustbaidu}.

However, the mainstream content moderation software is not robust enough~\cite{notrobustbaidu, notrobustfacebook}.
For example, Facebook content moderation software cannot understand many languages, leaving non-English speaking users more susceptible to harmful posts \cite{notrobustfacebook}.
In addition, toxic content can bypass mainstream content moderation software by applying simple textual transformations. For example, changing ``fuck'' to ``f$\mu$ck''.
The essential first step is to develop a testing framework for content moderation software to address this problem, similar to traditional software.

There remains a dearth of testing frameworks for content moderation software—partly because the problem is quite challenging.
First, most of the existing testing~\cite{He2020StructureInvariantTF,Sun2020AutomaticTA,Sun2022ImprovingMT} or adversarial attack~\cite{Li2020BERTATTACKAA, Garg2020BAEBA, Jin2020IsBR} techniques for Natural Language Processing (NLP) software rely on word-level semantic-preserving perturbations (e.g., from ``I \textit{like} it'' to ``I \textit{love} it'').
Most of the perturbed texts generated by these approaches still contain toxic words, and thus, they are unlikely to evade moderation.
In addition, as reported by a recent study~\cite{AEON2022ISSTA}, $44\%$ of the test cases generated by the State-of-the-Art (SOTA) approaches are false alarms, which are test cases with inconsistent semantics or incorrect grammar, rendering these approaches suboptimal.
Moreover, existing character-based perturbation approaches~\cite{Gao2018BlackBoxGO,Li2019TextBuggerGA,Boucher2022BadCI,Eger2019TextPL} are designed for general NLP software, so they consider common transformations (e.g., from ``foolish'' to ``folish''), which only cover a very limited set of the possible real user inputs for content moderation software.

In this paper, we propose \textit{\methodname}, a \underline{M}etamorphic \underline{T}esting framework for \underline{T}extual content \underline{M}oderation software.
Specifically, to develop a comprehensive testing framework for content moderation software, we first need to understand what kind of transformations real users might apply to evade moderation.
Thus, we conduct a pilot study (Section \ref{sec-mrs}) on $2,000$ text messages collected from real users and summarize eleven metamorphic relations across three perturbation levels: character level, word level, and sentence level, making \methodname provide metamorphic relations that reflect real-world user behaviors and are specially designed for content moderation software.
\methodname employs these metamorphic relations on toxic contents to generate test cases that are still toxic (i.e., being easily recognizable to humans) yet are likely to evade moderation.
All these metamorphic relations are implemented for two languages, English and Chinese, because English is a representative language based on the alphabet, while Chinese is a representative language based on the pictograph. 

We apply \methodname to test three commercial textual content moderation software and two SOTA moderation algorithms against three typical kinds of toxic content (i.e., abusive language, malicious advertisement, and pornography). 
The results show that \methodname achieves up to $83.9\%$, $51\%$, and $82.5\%$ error finding rates (EFR) when testing commercial content moderation software provided by Google, Baidu, and Huawei, respectively, and it obtains up to $91.2\%$ EFR when testing the SOTA algorithms from the academy.
In addition, we leverage the test cases generated by \methodname to retrain the model we explored, which largely improves model robustness ($0\%\sim5.9\%$ EFR) while maintaining the accuracy on the original test set. 
Codes, data and results of our pilot study in this paper are available\footnote{https://github.com/Jarviswang94/MTTM}.
The main contributions of this paper are as follows:
\begin{itemize}[leftmargin=*]
    \item The introduction of the first comprehensive testing framework, \methodname, for textual content moderation software validation.
    \item A pilot study on $2,000$ real-world text messages that leads to eleven metamorphic relations, facilitating the implementation of \methodname towards two languages: English and Chinese.
    \item An extensive evaluation of \methodname on three commercial content moderation software and two SOTA academic models,     demonstrating that \methodname can generate toxic contents that easily bypass moderation and those toxic contents can improve the robustness of the SOTA algorithms.
\end{itemize}

\noindent \textbf{Content Warning}: We apologize that this paper presents examples of aggressive, abusive, or pornographic expressions for clarity. Examples are quoted verbatim. In addition, to conduct this research safely, we performed the following precautionary actions for the participants: (1) in every stage, we prompted a content warning to the researchers and the annotators and told them that they could leave anytime during the study and (2) we provided psychological counseling after our study to relieve their mental stress.
\section{Background}
\label{sec-backgound}

\subsection{Textual Content Moderation}

\subsubsection{Commercial Content Moderation Software}

Many large companies, such as Google, Facebook, Twitter, and Baidu, have deployed commercial content moderation software on their products.
According to their official technical documents, the typical backbone of moderation software is a hybrid classification algorithm of neural network models and pre-defined rules, which leverages the advantages of both parties.
Neural network-based methods can effectively understand contextual and semantic information, while rule-based methods can easily implement user-defined functionality.
For example, Baidu's commercial content moderation software is powered by a deep neural network and a huge list of pre-defined banned words.

\subsubsection{Academic Content Moderation Models}

There are generally two categories of academic models for textual content moderation: \textit{feature engineering-based} models and \textit{neural network-based} models.

\noindent \textbf{Feature Engineering-Based Models}. Feature engineering-based models train their toxic content classification models based on manually-constructed features.
Specifically, textual feature engineering can be further divided into \textit{rule-based} methods and \textit{statistical} methods.

The core of rule-based methods is pre-defined rules or dictionaries of banned words.
Spertus et al. \cite{Spertus1997SmokeyAR} employed $47$ handcrafted linguistic rules to extract binary feature vectors and used a decision tree to detect toxic contents.
Razavi et al. \cite{Razavi2010AAI} constructed an abusive language dictionary to extract lexicon-level features for abuse detection.
Handcrafted rules and lexicons generalize well across data from different domains.
However, they can hardly deal with implicit abuse and sarcasm (e.g., ``I haven’t had an intelligent conversation with a woman
in my whole life'' from~\cite{Wiegand2021ImplicitlyAL}).
In addition, they are vulnerable to the detection of toxic text with errors in spelling, punctuation, and grammar~\cite{Wiegand2018InducingAL}. 

Statistical methods leverage different statistics of the textual data.
Yin et al. \cite{Yin2009DetectionOH} and Salminen et al. \cite{Salminen2018AnatomyOO} computed the Term Frequency-Inverse Document Frequency (TF-IDF) of words as features and used Support Vector Machines (SVMs) to detect online harassment and hate speech.
Statistical methods require less human effort and are more robust to spelling, punctuation, and grammar variations.
Nevertheless, these methods often capture superficial patterns instead of understanding the semantics~\cite{Wiegand2018InducingAL}.

\noindent \textbf{Neural Network-Based Models}. Advancements in text representation learning have spurred researchers to explore neural network-based models for textual content moderation.
Djuric et al. \cite{Djuric2015HateSD} was the first that utilized neural networks to obtain surface-level representations and trained a logistic regression classifier to detect abusive language.
Badjatiya et al. \cite{Badjatiya2017DeepLF} adopted GLoVe word embedding \cite{pennington2014glove} to extract text features and used a word-level LSTM to moderate textual content.
With the help of the pre-trained language models (e.g., BERT \cite{Devlin2019BERTPO} and RoBERTa \cite{Liu2019RoBERTaAR}), researchers fine-tune these models on a downstream dataset and achieved remarkable performance on textual content moderation tasks.

\subsection{Metamorphic Testing}


Metamorphic testing \cite{Chen2020MetamorphicTA} is a testing technique that has been widely employed to address the oracle problem.
The core idea of metamorphic testing is to detect violations of \textit{metamorphic relations} (MRs) across multiple runs of the software under test.
Specifically, MR describes the relationship between input-output pairs of software.
Given a test case, metamorphic testing transforms it into a new test case via a pre-defined transformation rule and then checks whether the corresponding outputs of these test cases returned by the software exhibit the expected relationship.

Metamorphic testing has been adapted to validate Artificial Intelligence (AI) software over the past few years.
These efforts aim to automatically report erroneous results returned by AI software via developing novel MRs.
In particular, Chen et al. \cite{Chen2008AnIA} investigated the use of metamorphic testing in bioinformatics applications.
Xie et al. \cite{Xie2011TestingAV} defined eleven MRs to test k-Nearest Neighbors and Naive Bayes algorithms. 
Dwarakanath et al. \cite{Dwarakanath2018IdentifyingIB} presented eight MRs to test SVM-based and ResNet-based image classifiers. 
Zhang et al. \cite{Zhang2018DeepRoadGM} tested autonomous driving systems by applying GANs to produce driving scenes with various weather conditions and checking the consistency of the system outputs.

\begin{table*}
\caption{Summary of the perturbation categories in the pilot study.}
\label{tab:mrs}
\centering
\begin{tabular}{l l l l l}
\toprule
\bf Perturbation Level & \bf Perturbation Method & \bf Examples in English & \bf Examples in Chinese & \bf Percentage\\
\midrule
\multirow{6}{*}{Character Level}
& Visual-based Substitution & a $\rightarrow$ $\alpha$; C $\rightarrow$ (; l $\rightarrow$ 1 & \begin{CJK}{UTF8}{gkai}日\end{CJK} $\rightarrow$ \begin{CJK}{UTF8}{gkai}曰\end{CJK}; \begin{CJK}{UTF8}{gkai}北\end{CJK} $\rightarrow$ \begin{CJK}{UTF8}{gkai}兆\end{CJK} & 12.3\% \\
& Visual-based Splitting & K $\rightarrow$ |<; W $\rightarrow$ VV & \begin{CJK}{UTF8}{gkai}好的\end{CJK} $\rightarrow$ \begin{CJK}{UTF8}{gkai}女子白勺\end{CJK}  & 5.0\%\\
& Visual-based Combination & Earn $\rightarrow$ Eam & \begin{CJK}{UTF8}{gkai}不用\end{CJK} $\rightarrow$ \begin{CJK}{UTF8}{gkai}甭\end{CJK} & 0.8\% \\
& Noise Injection & Hello $\rightarrow$ H**elll*o &  \begin{CJK}{UTF8}{gkai}致电\end{CJK} $\rightarrow$ \begin{CJK}{UTF8}{gkai}致\end{CJK}*\begin{CJK}{UTF8}{gkai}电\end{CJK} & 13.2\% \\
& Char Masking &  Hello $\rightarrow$ H*llo  & \begin{CJK}{UTF8}{gkai}新年快乐\end{CJK} $\rightarrow$  \begin{CJK}{UTF8}{gkai}新年快*\end{CJK} & 7.4\%\\
& Character Swap & Weather $\rightarrow$ Waether & \begin{CJK}{UTF8}{gkai}简单来说\end{CJK} $\rightarrow$ \begin{CJK}{UTF8}{gkai}简来单说\end{CJK} & 4.1\% \\
\midrule
\multirow{4}{*}{Word Level}
& Language Switch & Hello $\rightarrow$ Hola; + $\rightarrow$ Add & \begin{CJK}{UTF8}{gkai}龙\end{CJK} $\rightarrow$ \begin{CJK}{UTF8}{bkai}龍\end{CJK} & 14.9\%\\
& Homophone Substitution & Die $\rightarrow$ Dye; Night $\rightarrow$ Nite  & \begin{CJK}{UTF8}{gkai}好吧\end{CJK} $\rightarrow$ \begin{CJK}{UTF8}{gkai}猴八\end{CJK}; \begin{CJK}{UTF8}{gkai}这样\end{CJK} $\rightarrow$ \begin{CJK}{UTF8}{gkai}酱\end{CJK}  & 36.4\%\\
& Abbreviation Substitution & As Soon As Possible $\rightarrow$ ASAP & \begin{CJK}{UTF8}{gkai}永远的神\end{CJK} $\rightarrow$ yyds & 15.7\%\\
& Word Splitting & Hello $\rightarrow$ Hell o & \begin{CJK}{UTF8}{bkai}使用戶滿意\end{CJK} $\rightarrow$ \begin{CJK}{UTF8}{bkai}使用\end{CJK}..\begin{CJK}{UTF8}{bkai}戶滿意\end{CJK} & 6.6\%\\
\midrule
Sentence Level
& Benign Context Camouflage & \makecell[l]{Golden State Warriors guard won't \\ play Sunday, <add a spam sentence \\ here>, due to knee soreness.} & \makecell[l]{\begin{CJK}{UTF8}{gkai}金融业增加值超香港, <在这里\end{CJK} \\ \begin{CJK}{UTF8}{gkai}添加一条广告>, 是金融市场体系\end{CJK} \\ \begin{CJK}{UTF8}{gkai}最完备、集中度最高的区域。\end{CJK}} & 2.5\% \\
\bottomrule
\end{tabular}
\end{table*}
\section{\methodname}
\label{sec-mrs}

This section first introduces a pilot study on text messages collected from real users (Section \ref{sec:pilotstudy}).
Then we introduce eleven metamorphic relations that are inspired by the pilot study.
These metamorphic relations can be grouped into three categories according to the perturbation performed: character-level perturbations (Sec.~\ref{sec:charlevel}), word-level perturbations (Sec.~\ref{sec:wordlevel}), and sentence-level perturbations (Sec.~\ref{sec:sentencelevel}).

\subsection{Pilot Study}
\label{sec:pilotstudy}

In this work, we intend to develop metamorphic relations that assume the seed test case (i.e., a piece of text) and the perturbed test case should have identical classification labels (i.e., labeled as ``toxic content'') returned by the content moderation software. 
To generate effective test cases, we think the perturbations in our MRs should be:
\begin{itemize}[leftmargin=*]
    \item \textit{Semantic-preserving}: the perturbed test cases should have the identical semantic meaning as the seed.
    \item \textit{Realistic}: should reflect possible inputs from real users.
    \item \textit{Unambiguous}: should be defined clearly.
\end{itemize} 

In order to design satisfactory perturbations, we first conducted a pilot study on text messages from real users to explore what kind of perturbations the users would apply to the toxic content to bypass the content moderation software.
We consider text messages from four platforms with a large number of users:
\begin{itemize}[leftmargin=*]
    \item Twitter\footnote{https://twitter.com/} is a worldwide microblogging and social media platform on which users post and interact via messages known as ``tweets''.
    HateOffensive\footnote{https://github.com/t-davidson/hate-speech-and-offensive-language} \cite{hateoffensive} is a GitHub repository containing $24,802$ English hate speech sentences collected from Twitter. 
    \item Grumbletext\footnote{http://www.grumbletext.co.uk/} is a UK forum on which cell phone users make public claims about SMS spam messages.
    Kaggle released a spam classification competition dataset\footnote{https://www.kaggle.com/uciml/sms-spam-collection-dataset} with a collection of $5,574$ messages extracted manually from Grumbletext.
    \item Taobao\footnote{https://www.taobao.com/} is an e-commercial platform with around $900$ million active users.
    SpamMessage\footnote{https://github.com/hrwhisper/SpamMessage} is a dataset containing $10$ thousand user comments collected from Taobao.
    \item Dirty\footnote{https://github.com/pokemonchw/Dirty} is a GitHub repository containing $2,500$ Chinese toxic sentences with abusive and sexual words collected from Chinesse Internet community.
\end{itemize}

We randomly selected $2,000$ sentences from the above dataset for manual inspection and recruited three annotators to label all the sentences independently.
All the annotators have a Bachelor's degree or above and are proficient in both English and Chinese.
Annotators were given extensive guidelines, test tasks, and training sessions on content moderation software and toxic content.
For each sentence, annotators were asked two questions. (1) Whether the sentence is toxic or not? (2) Is the toxic content intentionally perturbed to bypass the content moderation software?
After the annotation, we use the label that most workers agree with as the final human label and finally obtain $1476$ toxic sentences with $121$ labeled as ``toxic and intentionally perturbed'' sentences.
We collected the contents labeled as toxic and intentionally perturbed by the annotators to design our perturbation methods.

We manually inspected all these toxic contents perturbed by the real users and collectively summarized eleven perturbation methods that real users have been using to evade moderation.
We categorize these toxic sentences from three perspectives: 1) basic unit of perturbation, such as character level, word level, and sentence level; 2) basic perturbation operation, such as substitution, insertion, deletion, split, and combination; and 3)
the logic behind perturbation, such as visual-based, homophone-based, and language-based.
Accordingly, we derive eleven MRs based on eleven perturbation methods, where each MR assumes that the classification label returned by the content moderation software on the generated test case (i.e., perturbed text) should be the same as that on the seed (i.e., original text).
Table~\ref{tab:mrs} presents the eleven perturbation methods, their categories, examples in two languages, and the percentage of each in our study.
We will introduce the MRs (their corresponding perturbation methods) in the following.

\subsection{MRs with Character-Level Perturbations}
\label{sec:charlevel}

\noindent \textbf{MR1-1 Visual-Based Substitution}

This MR uses visual-based substitutions, which replace characters with visually similar characters. 
These visually similar characters are not required to be semantically equivalent or similar to the original characters. 
Usually, the candidates come from the alphabet of other languages.
For example, users can replace ``a'' with ``å'', ``ä'', ``ą'', ``$\alpha$'', etc.
The candidates can also be punctuation or numbers, such as ``('' for ``C'' and ``1'' for ``l''.
For Chinese characters, we can consider their variants from different language systems, such as Kanji in Japanese, Hanja in Korean, and Han character in Vietnamese, making a Chinese character usually has up to three variants.
Besides variants, we can easily find many characters that look highly similar.
``\begin{CJK}{UTF8}{min}カ\end{CJK}'' (one of the Japanese kana) for ``\begin{CJK}{UTF8}{bkai}力\end{CJK}'' (Power) and ``\begin{CJK}{UTF8}{bkai}曰\end{CJK}'' (Say) for ``\begin{CJK}{UTF8}{bkai}日\end{CJK}'' (Sun) are examples of such substitutions.

\noindent \textbf{MR1-2 Visual-Based Splitting}

This MR employs visual-based splitting, which separates a character into multiple parts.
This MR is inspired by the fact that many characters are composed of other characters.
Therefore, some characters can be separated into two characters, such as ``VV'' for ``W'' and ``\begin{CJK}{UTF8}{gkai}女子\end{CJK}'' (Woman) for ``\begin{CJK}{UTF8}{gkai}好\end{CJK}'' (Good).
Some Chinese characters can even be split into three characters, for example ``\begin{CJK}{UTF8}{gkai}木身寸\end{CJK}'' (Wood/Body/Inch) for ``\begin{CJK}{UTF8}{gkai}榭\end{CJK}'' (Pavilion).
It is worth noting that Chinese characters can sometimes be split vertically, like ``\begin{CJK}{UTF8}{gkai}亡心\end{CJK}'' (Die/Heart) for ``\begin{CJK}{UTF8}{gkai}忘\end{CJK}'' (Forget).

\noindent \textbf{MR1-3 Visual-Based Combination}

This MR's perturbation method is the inverse transformation of MR1-2.
Visual-based combination combines adjacent characters into a single character, such as ``m'' for ``rn''.
The difference between this MR and MR1-2 is that, in MR1-2, the underlying meaning is expressed by the combination of characters. Instead, in this MR, we understand the meaning by splitting certain characters.

\noindent \textbf{MR1-4 Noise Injection}

This MR perturbs text via noise injection, which inserts additional characters into the original text.
To not affect human comprehension, users tend to let the noise be closely related to the context (e.g., ``o'' in ``Hellooo'') or from a different domain which can make users ignore the noise when reading (e.g., ``*'' in ``H*ell*o'').
Specifically, ``Hello'' has multiple ``o''s, and ``*'' is a mathematical symbol outside the English alphabet.
Therefore, humans can easily ignore the noises.

\noindent \textbf{MR1-5 Character Masking}

This MR uses character masking, which masks a small portion of the characters by replacing them with some special characters.
The content moderation software can hardly recognize the word, but humans can easily infer the masked character within the context.
For example, we can infer that the masked word is ``your'' in ``what's y*ur name'' with our prior knowledge.

\noindent \textbf{MR1-6 Character Swap}

``\textit{Aoccdrnig to a rscheearch at Cmabrigde Uinervtisy, it deosn't mttaer in waht oredr the ltteers in a wrod are, the olny iprmoetnt tihng is taht the frist and lsat ltteer be at the rghit pclae. The rset can be a toatl mses and you can sitll raed it wouthit porbelm. Tihs is bcuseae the huamn mnid deos not raed ervey lteter by istlef, but the wrod as a wlohe.}''\footnote{\url{https://www.mrc-cbu.cam.ac.uk/personal/matt.davis/Cmabrigde}}
Inspired by this fact, this MR uses character swap, which randomly swaps characters within a word.

\subsection{MRs with Word-Level Perturbations}
\label{sec:wordlevel}

\noindent \textbf{MR2-1 Language Switch}

This MR translates some words into other languages.
Many users on social media platforms can comprehend more than one language.
Thus, users may use words or phrases from different languages in a piece of text to evade moderation.  
Note that we also consider the switch between different written forms of a language as a language switch.
For example, in Chinese, it is commonly seen the transformation between traditional Chinese characters and simplified Chinese characters, such as ``\begin{CJK}{UTF8}{bkai}發\end{CJK}'' (Send) and ``\begin{CJK}{UTF8}{gkai}发\end{CJK}'' (Send).

\noindent \textbf{MR2-2 Homophone Substitution}

This MR is based on homophone substitution, which replaces words with other words or characters that have the same or similar pronunciation.
Simple examples include ``Dye'' ([da\i ]) for ``Die'' ([da\i ]), ``Nite'' ([na\i t]) for ``Night'' ([na\i t]) and ``C'' ([si:]) for ``see'' ([si:]).
Complex homophone substitution includes ``w8'' ([w] [e\i t])  for ``wait'' ([we\i t]), which uses a character outside English alphabet.

In Chinese, the pronunciation of ``\begin{CJK}{UTF8}{gkai}酱\end{CJK}'' ([t\textctc j\textscripta \textipa{N}], Sauce) is similar to that of ``\begin{CJK}{UTF8}{gkai}这样\end{CJK}'' ([t\textrtails \textgamma] [j\textscripta \textipa{N}], Such) when speaking fast.
In addition, the homophone class of a same character can vary in Chinese, leading to may possible substitutions.
For example, ``\begin{CJK}{UTF8}{gkai}重\end{CJK}'' (heterophones: [t\textrtails \textupsilon \textipa{N}], Repetition; or [t\textrtails $^h$\textupsilon \textipa{N}], Heavy) can be in the same homophone class with ``\begin{CJK}{UTF8}{gkai}虫\end{CJK}'' ([t\textrtails \textupsilon \textipa{N}], Insect), but it can be in the same homophone class with  ``\begin{CJK}{UTF8}{gkai}众\end{CJK}'' ([t\textrtails $^h$\textupsilon \textipa{N}], Many) as well.
Another example is that ``\begin{CJK}{UTF8}{gkai}九\end{CJK}'' (Nine) and ``\begin{CJK}{UTF8}{gkai}狗\end{CJK}'' (Dog) are in the same homophone class [k\textturna u] in Cantonese, but in different homophone class in Mandarin ([t\textctc jo\textipa{U}] and [ko\textipa{U}] respectively).

In addition, the substitution can happen between different languages. For example, ``exciting'' ([\i k\textprimstress sa\i t\i \textipa{N}]) and ``\begin{CJK}{UTF8}{gkai}亦可赛艇\end{CJK}'' ([\i] [k$^h$\textgamma][sa\i][t\i \textipa{N}], Also/Can/Race/Boat) are acoustically similar, and ``Bu'' is the Pinyin form of the Chinese character ``\begin{CJK}{UTF8}{gkai}不\end{CJK}'' ([pu], No).
Unlike the language switch in MR2-1, the perturbation logic behind this MR is homophone similarity rather than semantic equivalence.

\noindent \textbf{MR2-3 Abbreviation Substitution}

This MR focuses on abbreviation substitution.
Users tend to use the first letter to represent a word for convenience, such as ``ASAP'' for ``As Soon As Possible''.
In Chinese, people usually use the first letter of the characters' Pinyin to represent the characters.
For example, on social media platforms, ``YYDS'' is a common abbreviation for ``\begin{CJK}{UTF8}{gkai}永远的神\end{CJK}'' (Eternal God), whose Pinyin is ``Yong Yuan De Shen''.

\noindent \textbf{MR2-4 Word Splitting}

This MR injects spaces into the word, aiming to split a word into sub-words.
For example, ``Hello'' can be recognized in most popular NLP models.
If we add a space into the word, making it ``Hell o'', most NLP tokenizers will recognize it as two separate tokens, namely ``Hell'' and ``o'', which could affect the models' judgment.
This can also happen in Chinese.
For example, ``\begin{CJK}{UTF8}{gkai}使用户满意\end{CJK}'', which means ``satisfy the users'', should be tokenized as ``\begin{CJK}{UTF8}{gkai}使\end{CJK}/\begin{CJK}{UTF8}{gkai}用户\end{CJK}/\begin{CJK}{UTF8}{gkai}满意\end{CJK}''.
If we add some noises to separate the characters, it is easy to make the tokenization results become ``\begin{CJK}{UTF8}{gkai}使用\end{CJK}/\begin{CJK}{UTF8}{gkai}户\end{CJK}/\begin{CJK}{UTF8}{gkai}满意\end{CJK}'', which means ``Use/Household/Satisfy'', leading to the change of semantic meaning.

\subsection{MRs with Sentence-Level Perturbations}
\label{sec:sentencelevel}

\noindent \textbf{MR3-1 Benign Context Camouflage}

This MR uses benign context camouflage, which inserts plenty of benign or unrelated sentences to camouflage the toxic sentence.
For example, a malicious advertisement can be surrounded by numerous unrelated and non-commercial contents to bypass the malicious advertisement detection model.

\subsection{Discussion}

\noindent \textbf{Intersections of Different MRs.}
Some perturbations can fall into multiple MR categories.
For example, some substitution candidates not only have a similar visual appearance to the original character but also are the homophone of the original character, which corresponds to MR1-1 (visual-based substitution) and MR2-2 (homophone substitution), respectively.
In addition, similar-looking characters tend to have similar pronunciations, especially for Chinese.
However, the MR definitions are clear and can cover all the examples from our pilot study.
When counting the distribution, we randomly assign examples to one of the possible MRs.

\noindent \textbf{Combinations of Different MRs.}
We can use a combination of different MRs to generate diverse test cases.
However, to balance the generated test cases' diversity and readability, we restrict the maximum number of MRs used in each test case.
We evaluate the impact of MR combinations in Section \ref{subsec-testing-software-with-tool}.


\noindent \textbf{Generalization to other software and languages.}
In this work, we focus on textual content moderation software and implement our MRs for the two most widely used languages: English and Chinese.
However, based on our design methodology, these MRs can be easily generalized to other languages and to test other NLP software, such as software for user review analysis and machine translation.

\subsection{Implementation Details}

In this section, we describe the implementation details of \methodname.
Specifically, we implement (1) a target word selection approach and (2) the perturbations on the selected word in different MRs except MR3-1.
For MR3-1, we conduct sentence-level perturbation without the need to identify target words.

\noindent \textbf{Target Word Selection.}
We intend to perturb the words important for the content moderation scenario so that perturbations on these words are more likely to affect the output of content moderation software.
Specifically, we focus on words frequently appearing in the toxic content datasets but less frequently in a general domain corpus.
Thus, we use TF-IDF to select target words.
We utilize sklearn\footnote{https://scikit-learn.org/} for the English corpus and Jieba library\footnote{https://github.com/fxsjy/jieba} for the Chinese corpus.
After filtering out the stop words, we select the top $20$ words with the highest TF-IDF score for each dataset.

\noindent \textbf{MR1-1 Visual-Based Substitution.}
For each English character in the target words, we use DeepAI visual similarity API\footnote{https://deepai.org/machine-learning-model/image-similarity} to find the most visually similar character in the Greek and German alphabets as the candidate.
For each Chinese character in target words, we leverage SimilarCharacter\footnote{https://github.com/contr4l/SimilarCharacter}, a Python library that uses OpenCV\footnote{https://opencv.org/} to calculate the visual similarity score within $3,000$ commonly used Chinese characters, to find another word with the highest visual similarity score as the candidate.
To ensure a high similarity, we only replace the original character with the candidate if their similarity score is higher than $0.7$.

\noindent \textbf{MR1-2 Visual-Based Splitting.}
For both languages, we use DeepAI visual similarity API to find the most visually similar bi-char combinations as the candidate.
We only replace the original character with the candidate if their similarity score is higher than $0.7$.
Due to the large character space of Chinese characters, it is time-consuming to transverse all the bi-char combinations.
Thus, we use the Chinese Character Dictionaries\footnote{https://github.com/kfcd/chaizi} to split the character that is split-able in target words as the candidate.

\noindent \textbf{MR1-3 Visual-Based Combination.}
MR1-3 uses the splitting substitution (the original character, the candidate) dictionary built in MR1-2 (Visual-Based Splitting).
For each target word, if any of its bi-char combinations occur in the dictionary, we substitute the combined character for the bi-char combination.

\noindent \textbf{MR1-4 Noise Injection.}
We implement two character-level noise injection methods: insertion and repetition.
For insertion, we randomly insert a character into the target word.
According to the definition in Section \ref{sec:charlevel}, we implement two types of insertion: inserting a character from the language's alphabet, which is closely related to the context, and inserting a unique punctuation character, which is from a different domain.
For repetition, we repeat the vowel in each English target word and randomly repeat a character in each Chinese target word.

\noindent \textbf{MR1-5 Character Masking.}
For each target word, we randomly replace a character with ``*'' to mask the character.
For English, we mask a vowel in the target word.

\noindent \textbf{MR1-6 Character Swap.}
For each target word, we randomly swap two adjacent characters.
For Chinese, we randomly swap characters after tokenization.

\noindent \textbf{MR2-1 Language Switch.}
For each target word in English (\textit{resp}. Chinese), we invoke Google Translate API\footnote{https://translate.google.com/} to translate it into Spanish (\textit{resp}. English), which is the most widely used second language in the USA (\textit{resp}. China).

\noindent \textbf{MR2-2 Homophone Substitution}
We use the eng-to-ipa\footnote{https://github.com/mphilli/English-to-IPA} Python library to convert English words to International Phonetic Alphabet (IPA) and then find other English words with the most similar IPA as substitution candidates.
For Chinese, we use the pypinyin\footnote{https://github.com/mozillazg/python-pinyin} and pinyin2hanzi\footnote{https://github.com/letiantian/Pinyin2Hanzi} libraries to find the substitution candidates.

\noindent \textbf{MR2-3 Abbreviation Substitution.}
For English target words, we replace them with their acronym, which is the word composed of the first letters of the target words.
For Chinese target words, we first use the pypinyin Python library to convert them to Pinyin and then use the acronym of their Pinyin as the candidate.

\noindent \textbf{MR2-4 Word Splitting.}
For each target word, we randomly insert a blank space.

\noindent \textbf{MR3-1 Benign Context Camouflage.}
We randomly collect ten benign sentences for each dataset from its non-toxic class.
Then for each toxic sentence, we insert the benign sentence either before or after it.
\section{Evaluation}
\label{sec-experiment}

\begin{table}
\centering
\caption{Statistics of Toxic Datasets.}
\begin{tabular}{l r r r r}
\toprule
\bf Dataset & \bf \#Sent & \bf Lang &  \bf Type & \bf Source\\
\midrule
HateOffensive & 24.8K  &  English & Abuse   &  Twitter \\
Dirty  &  2.5K & Chinese & Abuse & Weibo \\
SMSSpam & 5.5k & English &  Spam & Grumbletext \\
SpamMessage & 60K & Chinese & Spam & Taobao\\
Sexting  & 0.5K & English & Porno & Github \\
Midu  & 7.3K & Chinese & Porno & Midu \\
\bottomrule
\end{tabular}
\label{tab:data-statistics}
\end{table}


To evaluate the effectiveness of \methodname, we use our method to test three commercial software products and two SOTA algorithms for content moderation.
In this section, we try to answer the following four Research Questions (RQs):
\begin{itemize}[leftmargin=*]
    \item RQ1: Are the test cases generated by \methodname toxic and realistic?
    \item RQ2: Can \methodname find erroneous outputs returned by content moderation software?
    \item RQ3: Can we utilize the test cases generated by \methodname to improve the performance of content moderation?
    \item RQ4: How would different factors affect the performance of \methodname?
\end{itemize}

\subsection{Experimental Settings}

\subsubsection{Datasets}

We used different kinds of datasets as seed data to validate \methodname.
Previous researchers have collected, labeled, and released various types of data for research purposes.
In this paper, we choose the datasets with the highest citations according to Google Scholar or those with the most stars on GitHub.
Other than the above-mentioned four datasets (in Section \ref{sec:pilotstudy}), namely HateOffensive, SMS Spam Collection, SpamMessage, and Dirty, we utilize another two datasets: Sexting\footnote{https://github.com/mathigatti/sexting-dataset}, an English pornographic text dataset containing $537$ sexual texting messages, and Midu \cite{Song2021EvidenceAN}, a Chinese novel paragraph dataset collected from an online literature reading platform called MiDu App\footnote{http://www.midureader.com/}, which is a corpus with $62,876$ paragraphs including $7,360$ pornographic paragraphs and $55,516$ normal paragraphs.
Important statistics of the six datasets are shown in Table~\ref{tab:data-statistics}.

\subsubsection{Software and Models Under Test}
\label{subsec:models}

We use \methodname to test commercial textual content moderation software products and SOTA academic models.
Commercial software products include Google Jigsaw’s Perspective\footnote{https://www.perspectiveapi.com/}, Baidu AI Cloud\footnote{https://ai.baidu.com/tech/textcensoring}, and Huawei Cloud\footnote{https://www.huaweicloud.com/product/textmoderation.html}.
These software products were tested against the three typical kinds of toxic content in our evaluation.
One exception is Google Jigsaw's moderation of malicious advertisements because Google does not provide such functionality.
They are all popular software products for content moderation developed by companies and can be accessed by registered users via their APIs.
For research models, we select models from GitHub and Huggingface Model Zoo\footnote{https://huggingface.co/models} with the highest downloads and stars in recent three years.
For abuse detection, we select HateXplain \cite{Mathew2021HateXplainAB}, a BERT model fine-tuned on abuse detection datasets.
For spam detection, we use a BERT model fine-tuned on the spam detection dataset, downloaded from Huggingface\footnote{https://huggingface.co/mrm8488/bert-tiny-finetuned-sms-spam-detection}.
Since there are no publicly available pornography detection models, we do not test this research model in our experiments.

\subsection{RQ1: Are the test cases generated by \methodname toxic and realistic?}

\methodname aims to generate test cases that are toxic and are as realistic as the ones real-world users produce to evade moderation.
Thus, in this section, we evaluate whether the generated test cases are still toxic (i.e., semantic-preserving) and whether they are realistic.
We generated $100$ sentences with each perturbation method (\textit{i.e.}, $1,100$ generated sentences in total) and recruited two annotators with Bachelor's degrees or above and proficiency in both English and Chinese.
After given guidelines and training sessions, the annotators were asked to annotate all the generated pairs, each containing an original and a perturbed sentence.
For each sentence pair, we asked the following two questions:
(1) From ``$1$ strongly disagree'' to ``$5$ strongly agree'', how much do you regard the sentence as toxic content (abuse, pornographic, or spam)? 
(2) From ``$1$ strongly disagree'' to ``$5$ strongly agree'', how much do you think the perturbation is realistic in the sense that real users may use it? 
Note that when asking whether a sentence is toxic or not, the original sentence and the perturbed sentence were not presented at the same time.
The annotators can only view one sentence each time from shuffled data when labeling the toxicity.
We would review test cases with any disagreement or unrealistic flags.
Annotation results show that the average toxic score is $4.51$, and the average realistic score is $4.12$.
We follow \cite{Kirk2021HatemojiAT} to measure the inter-rater agreement using Randolph’s Kappa, obtaining a value of $0.81$, which indicates ``almost perfect agreement''.

\begin{tcolorbox}[width=\linewidth, boxrule=0pt, colback=gray!20, colframe=gray!20]
\textbf{Answer to RQ1:}
The test cases generated by \methodname are toxic and realistic.
\end{tcolorbox}

\subsection{RQ2: Can \methodname find erroneous outputs returned by content moderation software?}
\label{subsec-testing-software-with-tool}

\begin{table}
\centering
\caption{Test Cases Statistic.}
\begin{tabular}{l c | r r}
\toprule
\bf Software & \bf Tasks & \bf Ori Num  &\bf Seed Num  \\
\midrule
\multirow{2}{*}{Google}& Abuse & 1,633  &  1,306\\
 & Porn   & 537  &  168 \\
\midrule
\multirow{3}{*}{Baidu} & Abuse & 1,515 & 985 \\
 & Porn &  258  &  153\\
& Spam & 1,000 & 280\\
\midrule
\multirow{3}{*}{Huawei}&Abuse  & 1,515 & 598\\
& Porn & 258 & 142\\
&Spam  & 1,000 & 288 \\
\midrule
\multirow{2}{*}{Academic Model} &  Abuse &1,633 & 659 \\
 &  Spam &746 & 674 \\
\bottomrule
\end{tabular}
\label{tab:test_case_stat}
\end{table}

\begin{table*}
\centering
\caption{Error Finding Rates of commercial content moderation software and Academic Models (AM).}
\begin{tabular}{l l  | l l l l | l l l | l l l}
\toprule
\multirow{2}{*}{\bf Level} & \multirow{2}{*}{\bf Perturbation Methods } & \multicolumn{4}{c}{\bf Abuse Detection}   & \multicolumn{3}{c}{\bf Spam Detection}  & \multicolumn{3}{c}{\bf Pron Detection}\\
\cmidrule(lr){3-6} \cmidrule(lr){7-9}  \cmidrule(lr){10-12}  
& &  \bf Google & \bf Baidu  &\bf Huawei & \bf AM  &  \bf Baidu  &\bf Huawei & \bf AM  & \bf Google & \bf Baidu  &\bf Huawei\\
\midrule
\multirow{6}{*}{Char}& Visual-based Substitution &   19.4 &28.0  & 75.9 &  91.2& 51.0 & 75.7 & 84.0  & 36.9 & 35.2 & 47.2 \\
& Visual-based Split  &  30.9 &  16.3 & 52.7 & 53.1 & 49.3 & 81.3 & 82.2  & 51.6 & 19.7 & 31.0 \\
& Noise Injection (non-lang) &  57.1  & 0.0 &2.2  & 88.9  & 0.0 & 1.8 & 28.8  & 9.2 & 0.0 & 0.4\\
& Noise Injection (lang)   &  72.7  & 12.1 & 56.2 & 88.9  & 49.3 & 63.5 & 79.2 & 19.5 & 19.7 & 49.3\\
& Char Masking  & 50.8 & 19.8 &  50.3 & 88.9  & 47.2 & 58.1 & 78.9 & 10.7 & 38.0 & 47.9\\
& Char Swap   &  64.3  & 10.2 & 54.8 & 66.2 & 47.5 & 55.6 & 75.7 & 23.0 & 18.1 & 46.5  \\
\midrule
\multirow{4}{*}{Word}& Language Switch  & 57.7  & 38.0 &  76.3 & 84.1 & 35.7 & 49.3 & 53.9 & 32.7 & 39.4 & 49.3\\
& Homophone Substitution  & 73.4  & 26.8 & 77.4 &  85.6  & 48.9 & 75.7 & 77.1 & 22.6 & 36.6 & 47.2\\
& Abbreviation Substitution  & 83.9  & 22.7 & 63.4 & 88.9  & 52.2 & 82.5 & 83.6 & 32.1 & 38.0 & 48.6\\
& Visual Split  &  68.2  & 0.0 & 0.0 &85.6   & 0.0 & 0.0 & 87.0 & 8.3 & 0.0 & 0.0\\
\midrule
Sentence &   Benign Context Camouflage &  41.7 &  24.7 &  0.0 & 4.6 & 8.5 & 0.0 & 0.0 & 50.0 & 42.4 & 0.0 \\
\midrule
Multi & Perturbation Combinations & 75.1 &  30.5 &  79.8 & 90.3  & 50.2 & 76.4 & 80.1 & 66.4 & 45.1 & 48.9\\
\bottomrule
\end{tabular}
\label{tab:abuse}
\end{table*}

\methodname aims to automatically generate test cases to find potential bugs in current content moderation software.
Hence, in this section, we evaluate the number of bugs that \methodname can find in the outputs of commercial content moderation software and academic models.
We first input all the original sentences and obtain the classification label for each software product or model under test.
If an original sentence was labeled as ``non-toxic'', it would be filtered out because we intend to find toxic contents that can evade moderation.
The remaining sentences will be regarded as seed sentences for test case generation.
The number of original sentences and seed sentences is presented in Table~\ref{tab:test_case_stat}. 
Then, we conduct perturbations in \methodname's MRs on the seed sentences to generate test cases. 
Finally, we use the generated test cases to validate the software products and academic models. 
In particular, we check whether these test cases were labeled as ``toxic'' or ``non-toxic''.
Since the generated text should preserve the semantics of the seed sentence, they are supposed to be labeled as ``toxic''. If not, the generated test cases evade the moderation of the software products or academic models, indicating erroneous outputs.
To evaluate how well \methodname does on generating test cases that trigger errors, we calculate Error Finding Rate (EFR), which is defined as follows:

$$\text{EFR} = \frac{\text{the number of misclassified test cases}}{\text{the number of generated test cases}} * 100\%.$$

The EFR results are shown in Table~\ref{tab:abuse}. 
In general, \methodname achieves high EFRs.
The EFRs of commercial software products are lower than that of academic models.
Using different MRs, MTTM achieves up to $83.9\%$, $51\%$, and $82.5\%$ EFR when testing moderation software provided by Google, Baidu, and Huawei, respectively, and it obtains up to $91.2\%$ EFR when testing the SOTA academic models.
We think it is because commercial software has been armed with various rule-based methods to detect input perturbation. For example, Baidu has a patent titled “Method and equipment for determining sensitivity of target text”\footnote{https://patents.google.com/patent/CN102184188A/en}. Specifically, they provide pre-service rules in their pretreatment unit to: 1) remove the unusual characters, such as “*”, “\%”, “\#”, “\$”, and 2) convert text strings with the deformed bodies, such as perpendicular shape literal and characters in a fancy style, to normal text strings.
Notably, all the academic models can detect sentence-level benign context camouflage, which may be due to the attention mechanism employed by these models.
In addition, all software products and models can pass the test cases generated on MR1-3 (Visual-Based Combination).
Therefore, we do not include the results in Tables~\ref{tab:abuse}. 
The performance of commercial textual content moderation software varies greatly against different kinds of toxic content.
For example, Google Jigsaw's Perspective performs much better on pornography detection than on abusive language detection.
It is probably because some abusive language, especially swear words like ``fuck'', is not taken that seriously on informal occasions.
The performance of Baidu AI Cloud on malicious advertisement detection is much worse than that on the other two tasks, which might be related to the fact that Baidu's revenue mainly comes from advertising.
In addition, there is a possible consensus among Chinese web users that malicious advertisement is not as bad as abusive language and pornography.
Therefore, companies seem to focus on different kinds of toxic content when developing their content moderation software.

As the biggest search engine company in China, the textual content moderation software in Baidu outperforms the one in Huawei, which is the biggest communication technology company in China.
It is probably because Baidu has more business scenarios to design more rules and collect more training data to improve content moderation software's performance.


\begin{tcolorbox}[width=\linewidth, boxrule=0pt, colback=gray!20, colframe=gray!20]
\textbf{Answer to RQ2:}
\methodname achieves up to $83.9\%$, $51\%$, and $82.5\%$ EFR when testing moderation software provided by Google, Baidu, and Huawei, respectively, and it obtains up to $91.2\%$ EFR when testing the SOTA academic models.
\end{tcolorbox}

\subsection{RQ3: Can we utilize the test cases generated by \methodname to improve the performance of content moderation?}

We have demonstrated that \methodname can generate toxic and realistic test cases that can evade the moderation of commercial software products and SOTA academic models.
As shown in the ``Abuse Detection'' column in Table~\ref{tab:abuse}, \methodname achieves high EFR on academic models for most of its MRs (e.g., $91.2\%$ for MR1-1 Visual-Based Substitution), indicating the generated test cases can easily fool the models.
The following substantial question is: can these test cases be utilized to improve the performance of content moderation?
In other words, we hope to improve model robustness.
A natural thought is to retrain the models using test cases generated by \methodname and check whether the retrained models are more robust to various perturbations.

Specifically, we select the Abuse Detection task and use the Hate-Offensive Dataset~\cite{hateoffensive}.
We split the dataset into three parts: training set, validation set, and test set with the ratio of $6$:$2$:$2$.
We first fine-tune a pre-trained BERT model~\cite{Devlin2019BERTPO} on the training set as our abuse detection model, which is a widely used scheme for text classification.
We adopt the default fine-tuning settings suggested by Huggingface\footnote{\url{https://huggingface.co/transformers/v3.2.0/custom_datasets.html}}.
Specifically, we train the model with $3$ epochs, a learning rate of $5\times 10^{-5}$, a batch size of $16$, $500$ warming up steps, and a weight decay of $0.01$. 
We select the model with the highest accuracy on the validation set and use \methodname to test its robustness.

Then, for retraining with \methodname, we conduct fine-tuning with the failed test cases generated by \methodname.
We generated test cases with \methodname and randomly collected $300$ cases that could fool the model.
Labeling them as toxic contents, we add them to the original training set to retrain the model.
The setting of hyper-parameters is identical to that of regular training mentioned above.



\begin{table}
\centering
\caption{Error Finding Rates (EFRs) on abusive language detection models after retraining on the original test set and the test cases generated by \methodname.}
\centering
\begin{tabular}{l l l | l }
\toprule
\bf Level & \bf Perturb Methods & \bf Ori  &\bf Aug  \\
\midrule
\multirow{4}{*}{Char}& Visual-Based Substitution &  71.3  & 0.0 \\
& Visual-Based Splitting & 49.5 & 1.4 \\
& Noise Injection (non-lang) &  56.1  & 2.5\\
& Noise Injection (lang) & 56.1 & 2.5 \\
& Char Masking &  43.9  & 2.5 \\
& Char Swap & 45.6 & 3.0 \\
\midrule
\multirow{4}{*}{Word}& Language Switch & 76.2  & 5.9\\
& Homophone Substitution & 62.5 & 3.1\\
& Abbreviation Substitution & 76.2 & 2.2 \\
& Visual Splitting & 71.3  & 2.0 \\
\midrule
Sentence &   Benign Context Camouflage   &12.0 & 0.0 \\
\midrule
Multi & Perturbation Combinations &  81.4 & 3.5\\
\bottomrule
\end{tabular}
\label{tab:abuse_improve}
\end{table}

To validate the effectiveness of robust retraining with \methodname, we use \methodname to test the model after robust retraining, denoted as ``Aug'', and compared the EFRs with the original model's, denoted as ``Ori''.
The results are presented in Table~\ref{tab:abuse_improve}.
We can observe that the test case generated by \methodname can largely improve the robustness of the content moderation models in the sense that the EFRs have been significantly reduced (e.g., from $71.3\%$ to $0.0\%$ for the MR1-1 Visual-Based Substitution).
In other words, after retraining with \methodname's test cases, the model is rarely fooled by all the perturbations.
Moreover, the model's accuracy remains on par after robust training  (from 91.5\% to 91.2 \%), which means the retraining did not affect model performance on the original test set.

Notably, our approach will not introduce extra unknown tokens because: (1) BERT has a huge ($\sim 30,000$ tokens) vocabulary generated from massive data on the web, including characters from various languages; (2) BERT uses byte-pair encoding, an encoding technique that can effectively mitigate the out-of-vocabulary problem.
For example, the generated ``helllo'' will be tokenized into ``hell'' and ``lo'' instead of treating the whole word as an unknown token.

We do not conduct experiments on improving industrial models because industrial moderation only provides APIs while robust retraining requires access to model internals. However, we believe robust retraining with MTTM’s test cases would also improve the robustness of industrial models because the underlying models are similar. In the future, we can study on how to improve the robustness of industrial moderation by designing a preprocessing module to detect and filter out/reverse-perturb intentionally-perturbed inputs.

\begin{figure}
\centering
\includegraphics[width=0.48\textwidth]{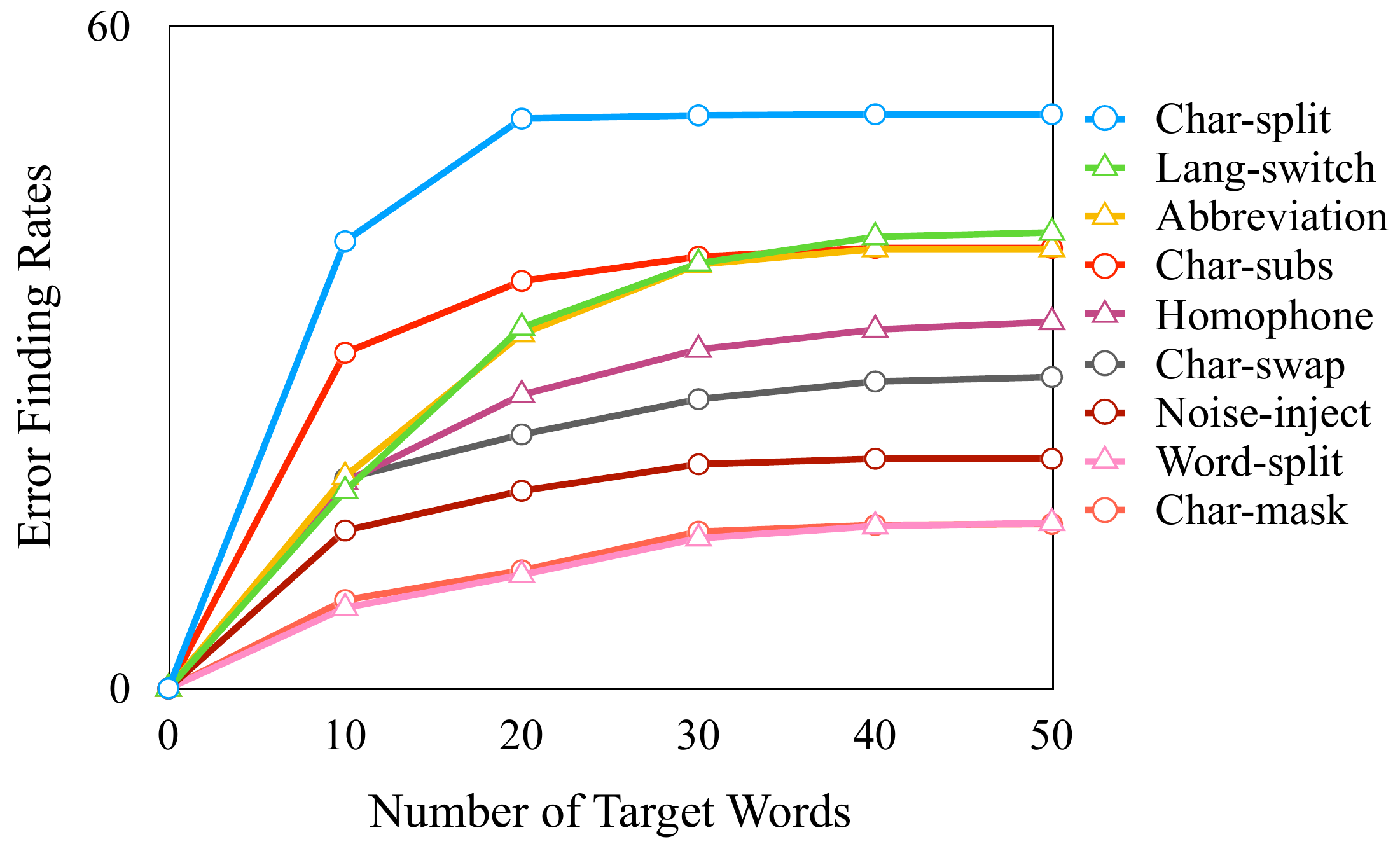}
\caption{The Errors Finding Rates of \methodname with different number of target words.}
\label{fig:num_target}
\end{figure}
\begin{tcolorbox}[width=\linewidth, boxrule=0pt, colback=gray!20, colframe=gray!20]
\textbf{Answer to RQ3:}
Test cases generated by \methodname can effectively improve the robustness of academic content moderation models.
\end{tcolorbox}

\subsection{RQ4: How would different factors affect the performance of \methodname?}

This section explores the impact of four factors on the performance of \methodname.
First, we studied the impact of noisy character selection on the performance of our method.
In the previous sections, we observe that inserting noisy characters into target words (MR1-4) can help bypass the content moderation software and models.
To study the impact of noisy character selection, we try two types of noisy characters: characters from the dataset and special characters that are not in the dataset.
As shown in Table~\ref{tab:abuse}, inserting characters from the dataset as noise (dubbed Noise Injection (lang)) is much more effective than inserting special characters that are not in the dataset (named Noise Injection (non-lang)).
One possible reason is that commercial software products have designed some rule-based preprocessing to the input sentence to remove special tokens that are not commonly seen or recover non-English characters (\textit{e.g.}, ä) to English characters (\textit{e.g.}, a).
These techniques are usually called text normalization.

Second, we studied the impact of the number of target words.
We calculated the TF-IDF scores in the previous sections and selected the top $20$ words as target words.
To study the impact of the number of target words, we vary the number of target words from $10$ to $50$ and compute the corresponding EFRs.
As shown in Fig.~\ref{fig:num_target}, \methodname can find more errors as the number of target words increases.
However, the EFRs saturate when the number of target words is larger than $40$.

\begin{figure}
\centering
\includegraphics[width=0.4\textwidth]{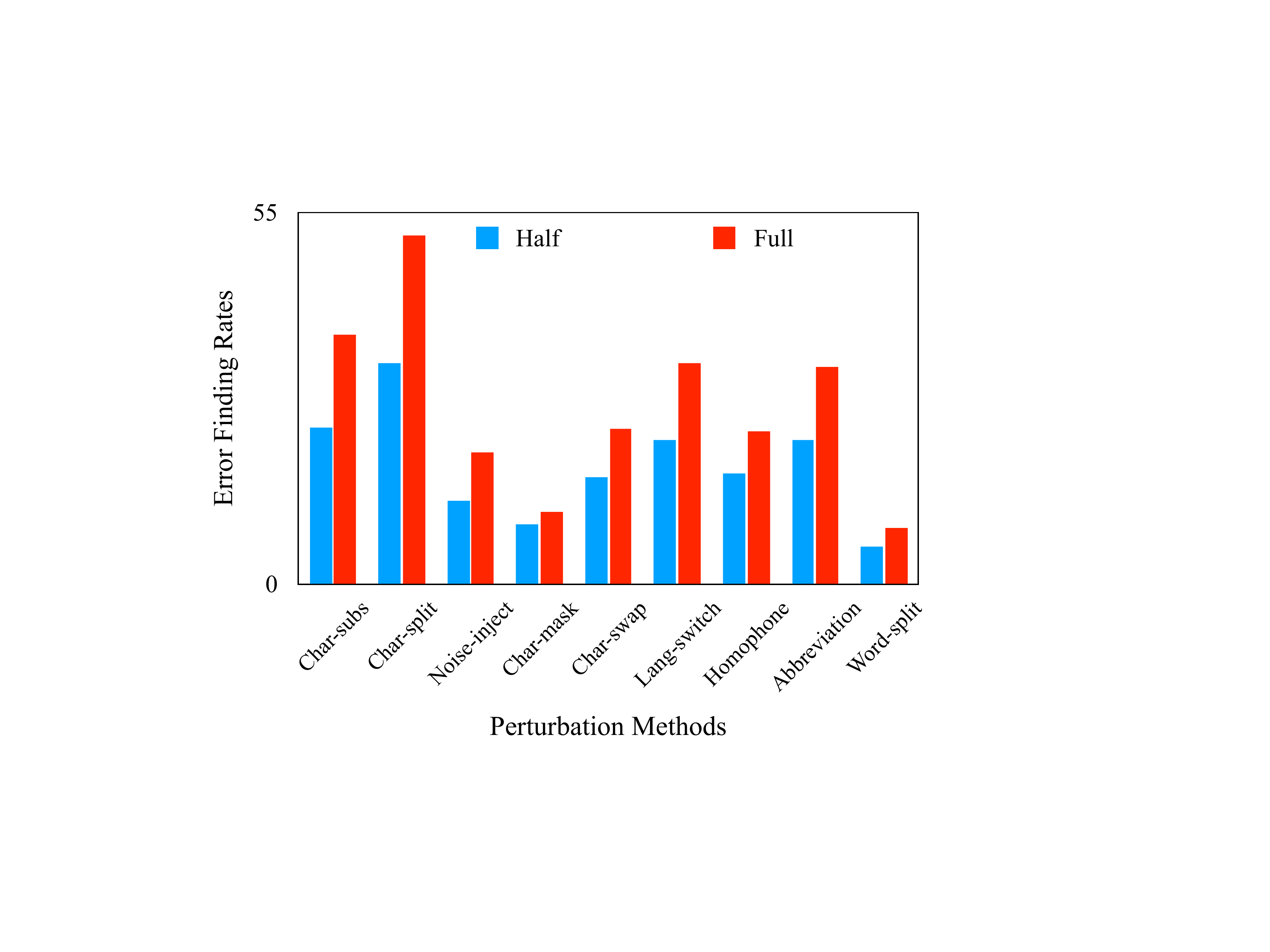}
\caption{The Error Finding Rates of different perturbation numbers to be applied to a single example.}
\label{fig:num_purturb}
\end{figure}

Third, we studied the impact of the number of perturbations.
In the previous sections, we perturbed all the target words in each sentence.
In this experiment, for each sentence, we compare the EFRs of perturbing all the target words and that of randomly perturbing half of the target words.
As shown in Fig.~\ref{fig:num_purturb}, perturbing all the target words in each sentence can significantly improve the EFRs.
Only perturbing half of the target words in each sentence is not sufficient to bypass the content moderation software.

Last but not least, we studied the impact of the perturbation combinations.
In the previous sections, we showed that using each perturbation method alone can achieve a good EFR.
To study the impact of different perturbation combinations, we randomly select one char-level perturbation and one word-level perturbation, leading to $24$ ($6\times 4$) combinations.
According to the results in Table~\ref{tab:abuse}, combining the different perturbation levels can increase the EFR.

\begin{tcolorbox}[width=\linewidth, boxrule=0pt, colback=gray!20, colframe=gray!20]
\textbf{Answer to RQ4:}
Noisy characters from the same dataset, more target words, more perturbations, and the combination of different perturbations can boost the performance of \methodname.
\end{tcolorbox}

\subsection{Compared with Textual Adversarial Attack Methods}

In this section, we will illustrate the advantage of \methodname compared to textual adversarial attack methods, which is another line of research for finding the error in NLP software.

First, \methodname is more comprehensive than adversarial methods because most of these methods focus on a small subset of the perturbations in MTTM. In addition, as reported by recent studies~\cite{Morris2020ReevaluatingAE, AEON2022ISSTA}, textual adversarial attack methods often generate low-quality test cases because their semantics change in many cases (around 40\%), while \methodname can generate toxic and realisitic test cases (Section IV B).

To show the effectiveness of \methodname, we conduct an expeirment to compare the performance of \methodname with textual adversarial attacks methods in terms of EFR and running time. Specifically, we attacked our BERT-based abusive detection model in English using two famous NLP adversarial methods: PSO~\cite{Zang2019WordlevelTA} and BAE~\cite{Garg2020BAEBA}, leading to an EFR of 65.0\% and 47.8\%, respectively, while a majority of MTTM’s MRs achieve more than 85\% EFR (Table IV). In addition, adversarial methods need much more running time than MTTM because these methods rely on extensive model queries, while MTTM needs one query per test case. The running time of the two adversarial methods are 605.2x and 72.5x more. In summary, \methodname can find more error in less running time.
\section{Threats to Validity}
\label{sec-discuss}


The validity of our study may be subject to some threats.
\textit{The first threat} is that the test cases generated by \methodname after many perturbations may become ``non-toxic'', leading to false positives.
To relieve this threat, we conducted a user study to validate whether the generated test cases are toxic or not.
We further asked the annotators to label whether the test cases reflect inputs from real users. The results show that the generated test cases are toxic and realistic.
\textit{The second threat} is that we implement \methodname for two languages, which may not generalize to other natural languages.
To reduce this threat, the choice of the two languages is made thoughtfully: they are representative alphabet-based language and pictograph-based language, respectively.
In addition, we believe our MRs can generalize to other languages because most of the languages share similar properties (e.g., visual similarity, homophone, language switch).
\textit{The third threat} lies in our evaluation of five content moderation systems, which might not be a proper estimate of \methodname's performance on other systems.
We test commercial content moderation software and SOTA academic models to mitigate this threat.
In particular, we test content moderation software provided by three big companies, which already have their techniques to defend malicious inputs.
In the future, we could test more commercial software and research models to further mitigate this threat.
\textit{The fourth threat} is that our \methodname could be outdated with the bypass techniques evolving.
To reduce this threat, we provide a comprehensive workflow: study the user behaviors, summarize and design the MRs, generate test cases, and use failure cases to improve the robustness.
If other bypass techniques were proposed, people could follow this workflow to design new MRs.
We also believe that automated MR generation is a promising and useful direction. This line of research mainly focuses on automated generation of a specific kind of MRs (e.g., polynomial MRs~\cite{Zhang2014SearchbasedIO, Zhang2019AutomaticDA} or automated MR generation leveraging software redundancy~\cite{Carzaniga2014CrosscheckingOF}. Since automated MR generation for content moderation software faces different challenges, we regard it as an important future work.



\section{Related Work}

\subsection{Robustness of AI Software}

\textit{AI software} has been adopted by various domains, such as autonomous driving and face recognition.
However, AI software is not robust enough and can generate erroneous outputs that lead to fatal accidents \cite{notrobustself-driving,notrobusttesla}.
To this end, researchers have proposed a variety of methods to generate adversarial examples or test cases that can fool AI software~\cite{Carlini2016HiddenVC, Tu2021ExploringAR, Luo2021InteractivePF, Pei2017DeepXploreAW, Zhang2022MachineLT,Riccio2020TestingML,Humbatova2021DeepCrimeMT,Pham2021DEVIATEAD,Wang2021RobOTRT,Huang2021CoverageGuidedTF,Zhang2022ImprovingAT}.
Meanwhile, researchers have also designed approaches to improve AI software’s robustness, for example, the robust training mechanism~\cite{Madry2018TowardsDL, Asyrofi2021CanDT, Gao2020FuzzTB} and network debugging~\cite{Ma2018MODEAN,Tao2020TRADERTD}.
\textit{NLP software} has also been used in recent years.
Typical scenarios include sentiment analysis~\cite{Zhang2017SentimentAA,Wang2017EmotionRW}, machine translation~\cite{Bahdanau2015NeuralMT,Wang2022UnderstandingAI,Jiao2022TencentsMM} and text-to-speech synthesis~\cite{Wang2017TacotronTE, Ma2018FPETSFP}. 
Because of its importance, researchers from both NLP and software engineering areas have started to explore the robustness of NLP software~\cite{Gupta2020MachineTT, He2021TestingMT, Jiao2023IsCA}.
In particular, Ribeiro et al.~\cite{Ribeiro2020BeyondAB} designed a behavioral testing method to test NLP software for sentiment analysis, duplicate question answering, and machine comprehension.
Li et al.~\cite{Li2020BERTATTACKAA} used deep learning models to generate test cases for deep learning-based NLP software.
Sun et al.~\cite{Sun2022ImprovingMT} propose a word-replacement-based approach to test and fix machine translation bugs without re-training.
Our paper studies the robustness of a widely-used AI software, content moderation software, which has not been systematically studied.

\subsection{Robustness of Textual Content Moderation Software}

We systematically reviewed papers on testing and attacking textual content moderation across related research areas: software engineering, natural language processing, and speech signal processing. 
Specifically, Ahlgren~\cite{Ahlgren2021TestingWE} used metamorphic testing to test Facebook's simulation system, which is used to tackle harmful content.
Li et al.~\cite{Li2019TextBuggerGA} reported that visual-based substitution (MR1-1), character swap (MR1-6), and word splitting (MR2-4) could fool the NLP model.
Gao et al.~\cite{Gao2018BlackBoxGO} proposed a black-box attack method based on character swap (MR1-6) to fool deep learning classifiers.
Eger et al.~\cite{Eger2019TextPL} use visual-based substitution (MR1-1) to attack NLP models.
Kapoor et al.~\cite{Kapoor2019MindYL} stated that Indian Internet users could use English-Hindi code-switched language to express abusive content (MR2-1).
Cid et al.~\cite{Cid2008TheIO} found that spammers reduce the effectiveness of the spam detection algorithm by introducing noise in their messages (MR1-4).
Li et al.~\cite{Li2021EnhancingMR} found that malicious Chinese netizens may obfuscate some toxic words in their comments with the corresponding variants that are visually similar to the original words (MR1-1). 

However, our paper has sufficient contribution compared with the above papers.
First, \methodname is much more comprehensive.
Only five kinds of perturbations explored in these papers overlap with our MRs.
To the best of our knowledge, the other six MRs in \methodname have not been explored in the existing papers across different related research areas.
Moreover, all these papers focus on one language setting, while we implement \methodname for both English and Chinese.
In addition, all the MRs were supported by our pilot study on real user inputs, which are different from existing papers that came up with the perturbations based on domain knowledge.
Furthermore, most of the existing papers were only evaluated on research models, while \methodname has also been evaluated on three commercial content moderation software products.
Thus, we believe \methodname is the first comprehensive testing framework for textual content moderation.

\section{Conclusion}

This paper proposed the first comprehensive testing framework \methodname for validating textual content moderation software.
Unlike existing testing or adversarial attack technique for general NLP software, which only provide common perturbations and cover a very limited set of toxic inputs that malicious users may produce, \methodname contains eleven metamorphic relations that are mainly inspired by a pilot study. 
In addition, all the metamorphic relations in \methodname have been implemented for two languages: English and Chinese.
Our evaluation shows that the test cases generated by \methodname can easily evade the moderation of two SOTA moderation algorithms and commercial content moderation software provided by Google, Baidu, and Huawei.
The test cases have been utilized to retrain the algorithms, which exhibited substantial improvement in model robustness while maintaining identical accuracy on the original test set.
We believe that this work is the crucial first step toward systematic testing of content moderation software.
For future work, we will continue developing metamorphic relations in \methodname and extend it to more language settings.
We will also launch an extensive effort to help continuously test and improve content moderation software.

\section{Acknowledgement}
The work described in this paper was supported by the Research Grants Council of the Hong Kong Special Administrative Region, China (No. CUHK 14206921 of the General Research Fund) and the National Natural Science Foundation of China (Grant Nos. 62102340 and 62206318).

\balance
\bibliographystyle{IEEEtran}
\bibliography{reference}

\begin{thebibliography}{10}
\providecommand{\url}[1]{#1}
\csname url@samestyle\endcsname
\providecommand{\newblock}{\relax}
\providecommand{\bibinfo}[2]{#2}
\providecommand{\BIBentrySTDinterwordspacing}{\spaceskip=0pt\relax}
\providecommand{\BIBentryALTinterwordstretchfactor}{4}
\providecommand{\BIBentryALTinterwordspacing}{\spaceskip=\fontdimen2\font plus
\BIBentryALTinterwordstretchfactor\fontdimen3\font minus
  \fontdimen4\font\relax}
\providecommand{\BIBforeignlanguage}[2]{{%
\expandafter\ifx\csname l@#1\endcsname\relax
\typeout{** WARNING: IEEEtran.bst: No hyphenation pattern has been}%
\typeout{** loaded for the language `#1'. Using the pattern for}%
\typeout{** the default language instead.}%
\else
\language=\csname l@#1\endcsname
\fi
#2}}
\providecommand{\BIBdecl}{\relax}
\BIBdecl

\bibitem{tweets2020}
D.~Sayce, ``The number of tweets per day in 2020,''
  \url{https://www.dsayce.com/social-media/tweets-day/}, 2020, accessed:
  2022-03-01.

\bibitem{Badjatiya2017DeepLF}
P.~Badjatiya, S.~Gupta, M.~Gupta, and V.~Varma, ``Deep learning for hate speech
  detection in tweets,'' \emph{Proceedings of the 26th International Conference
  on World Wide Web Companion}, 2017.

\bibitem{Li2012KnowingYE}
Z.~Li, K.~Zhang, Y.~Xie, F.~Yu, and X.~Wang, ``Knowing your enemy:
  understanding and detecting malicious web advertising,'' \emph{Proceedings of
  the 2012 ACM conference on Computer and communications security}, 2012.

\bibitem{Rowley2006LargeSI}
H.~A. Rowley, Y.~Jing, and S.~Baluja, ``Large scale image-based adult-content
  filtering,'' in \emph{VISAPP}, 2006.

\bibitem{children2011}
E.~R. Munro, ``The protection of children online: a brief scoping review to
  identify vulnerable groups,'' \emph{Childhood Wellbeing Research Centre},
  2011.

\bibitem{spam2022}
N.~Cveticanin, ``What's on the other side of your inbox - 20 spam statistics
  for 2022,'' \url{https://dataprot.net/statistics/spam-statistics/}, 2022,
  accessed: 2022-03-01.

\bibitem{Yu2016InternetMI}
T.-K. Yu and C.-M. Chao, ``Internet misconduct impact adolescent mental health
  in taiwan: The moderating roles of internet addiction,'' \emph{International
  Journal of Mental Health and Addiction}, vol.~14, pp. 921--936, 2016.

\bibitem{Chen2020AutomaticDO}
Y.~Chen, R.~Zheng, A.~Zhou, S.~Liao, and L.~Liu, ``Automatic detection of
  pornographic and gambling websites based on visual and textual content using
  a decision mechanism,'' \emph{Sensors (Basel, Switzerland)}, vol.~20, 2020.

\bibitem{Mishra2019TacklingOA}
P.~Mishra, H.~Yannakoudakis, and E.~Shutova, ``Tackling online abuse: A survey
  of automated abuse detection methods,'' \emph{ArXiv}, vol. abs/1908.06024,
  2019.

\bibitem{Schmidt2017ASO}
A.~Schmidt and M.~Wiegand, ``A survey on hate speech detection using natural
  language processing,'' in \emph{SocialNLP@EACL}, 2017.

\bibitem{Wu2018TwitterSD}
T.~Wu, S.~Wen, Y.~Xiang, and W.~Zhou, ``Twitter spam detection: Survey of new
  approaches and comparative study,'' \emph{Comput. Secur.}, vol.~76, pp.
  265--284, 2018.

\bibitem{Devlin2019BERTPO}
J.~Devlin, M.-W. Chang, K.~Lee, and K.~Toutanova, ``Bert: Pre-training of deep
  bidirectional transformers for language understanding,'' \emph{NAACL}, vol.
  abs/1810.04805, 2019.

\bibitem{Liu2019RoBERTaAR}
Y.~Liu, M.~Ott, N.~Goyal, J.~Du, M.~Joshi, D.~Chen, O.~Levy, M.~Lewis,
  L.~Zettlemoyer, and V.~Stoyanov, ``Roberta: A robustly optimized bert
  pretraining approach,'' \emph{ArXiv}, vol. abs/1907.11692, 2019.

\bibitem{google2021}
L.~Hanu, J.~Thewlis, and S.~Haco, ``How ai is learning to identify toxic online
  content,''
  \url{https://www.scientificamerican.com/article/can-ai-identify-toxic-online-content/},
  2021, accessed: 2022-03-01.

\bibitem{facebook2020}
J.~Vincent, ``Facebook is now using ai to sort content for quicker
  moderation,''
  \url{https://www.theverge.com/2020/11/13/21562596/facebook-ai-moderation},
  2020, accessed: 2022-03-01.

\bibitem{twitter2020}
T.~Gillespie, ``Content moderation, ai, and the question of scale,'' \emph{Big
  Data \& Society}, vol.~7, no.~2, p. 2053951720943234, 2020.

\bibitem{notrobustbaidu}
M.~Jing, ``China’s baidu turns to ai to police online content, but is the
  technology reliable?''
  \url{https://www.scmp.com/tech/innovation/article/2143759/chinas-baidu-turns-ai-police-online-content-technology-reliable?module=perpetual_scroll_0&pgtype=article&campaign=2143759},
  2018, accessed: 2022-03-01.

\bibitem{notrobustfacebook}
K.~Canales, ``Facebook's ai moderation reportedly can't interpret many
  languages, leaving users in some countries more susceptible to harmful
  posts,''
  \url{https://www.businessinsider.com/facebook-content-moderation-ai-cant-speak-all-languages-2021-9},
  2021, accessed: 2022-03-01.

\bibitem{He2020StructureInvariantTF}
P.~He, C.~Meister, and Z.~Su, ``Structure-invariant testing for machine
  translation,'' \emph{2020 IEEE/ACM 42nd International Conference on Software
  Engineering (ICSE)}, pp. 961--973, 2020.

\bibitem{Sun2020AutomaticTA}
Z.~Sun, J.~M. Zhang, M.~Harman, M.~Papadakis, and L.~Zhang, ``Automatic testing
  and improvement of machine translation,'' \emph{2020 IEEE/ACM 42nd
  International Conference on Software Engineering (ICSE)}, pp. 974--985, 2020.

\bibitem{Sun2022ImprovingMT}
Z.~Sun, J.~Zhang, Y.~Xiong, M.~Harman, M.~Papadakis, and L.~Zhang, ``Improving
  machine translation systems via isotopic replacement,'' \emph{2022 IEEE/ACM
  44th International Conference on Software Engineering (ICSE)}, pp.
  1181--1192, 2022.

\bibitem{Li2020BERTATTACKAA}
L.~Li, R.~Ma, Q.~Guo, X.~Xue, and X.~Qiu, ``Bert-attack: Adversarial attack
  against bert using bert,'' \emph{EMNLP}, vol. abs/2004.09984, 2020.

\bibitem{Garg2020BAEBA}
S.~Garg and G.~Ramakrishnan, ``Bae: Bert-based adversarial examples for text
  classification,'' \emph{EMNLP}, 2020.

\bibitem{Jin2020IsBR}
D.~Jin, Z.~Jin, J.~T. Zhou, and P.~Szolovits, ``Is bert really robust? a strong
  baseline for natural language attack on text classification and entailment,''
  in \emph{AAAI}, 2020.

\bibitem{AEON2022ISSTA}
J.~Huang, J.~Zhang, W.~Wang, P.~He, Y.~Su, and M.~R. Lyu, ``{AEON:} a method
  for automatic evaluation of {NLP} test cases,'' in \emph{International
  Symposium on Software Testing and Analysis (ISSTA)}, 2022.

\bibitem{Gao2018BlackBoxGO}
J.~Gao, J.~Lanchantin, M.~L. Soffa, and Y.~Qi, ``Black-box generation of
  adversarial text sequences to evade deep learning classifiers,'' \emph{2018
  IEEE Security and Privacy Workshops (SPW)}, pp. 50--56, 2018.

\bibitem{Li2019TextBuggerGA}
J.~Li, S.~Ji, T.~Du, B.~Li, and T.~Wang, ``Textbugger: Generating adversarial
  text against real-world applications,'' \emph{NDSS}, 2019.

\bibitem{Boucher2022BadCI}
N.~P. Boucher, I.~Shumailov, R.~Anderson, and N.~Papernot, ``Bad characters:
  Imperceptible nlp attacks,'' \emph{2022 IEEE Symposium on Security and
  Privacy (SP)}, pp. 1987--2004, 2022.

\bibitem{Eger2019TextPL}
S.~Eger, G.~G. Sahin, A.~R{\"u}ckl{\'e}, J.-U. Lee, C.~Schulz, M.~Mesgar,
  K.~Swarnkar, E.~Simpson, and I.~Gurevych, ``Text processing like humans do:
  Visually attacking and shielding nlp systems,'' \emph{NAACL}, 2019.

\bibitem{Spertus1997SmokeyAR}
E.~Spertus, ``Smokey: Automatic recognition of hostile messages,'' in
  \emph{AAAI/IAAI}, 1997.

\bibitem{Razavi2010AAI}
A.~H. Razavi, D.~Inkpen, S.~Uritsky, and S.~Matwin, ``Offensive language
  detection using multi-level classification,'' in \emph{Advances in Artificial
  Intelligence}, 2010.

\bibitem{Wiegand2021ImplicitlyAL}
M.~Wiegand, J.~Ruppenhofer, and E.~Eder, ``Implicitly abusive language – what
  does it actually look like and why are we not getting there?'' in
  \emph{NAACL}, 2021.

\bibitem{Wiegand2018InducingAL}
M.~Wiegand, J.~Ruppenhofer, A.~Schmidt, and C.~Greenberg, ``Inducing a lexicon
  of abusive words – a feature-based approach,'' in \emph{NAACL}, 2018.

\bibitem{Yin2009DetectionOH}
D.~Yin, Z.~Xue, L.~Hong, B.~D. Davison, A.~Kontostathis, and L.~Edwards,
  ``Detection of harassment on web 2.0,'' \emph{Proceedings of the Content
  Analysis in the WEB}, vol.~2, pp. 1--7, 2009.

\bibitem{Salminen2018AnatomyOO}
J.~O. Salminen, H.~Almerekhi, M.~Milenkovic, S.-G. Jung, J.~An, H.~Kwak, and
  B.~J. Jansen, ``Anatomy of online hate: Developing a taxonomy and machine
  learning models for identifying and classifying hate in online news media,''
  in \emph{ICWSM}, 2018.

\bibitem{Djuric2015HateSD}
N.~Djuric, J.~Zhou, R.~Morris, M.~Grbovic, V.~Radosavljevic, and N.~L.
  Bhamidipati, ``Hate speech detection with comment embeddings,''
  \emph{Proceedings of the 24th International Conference on World Wide Web},
  2015.

\bibitem{pennington2014glove}
J.~Pennington, R.~Socher, and C.~D. Manning, ``Glove: Global vectors for word
  representation,'' in \emph{Proceedings of the 2014 conference on empirical
  methods in natural language processing (EMNLP)}, 2014, pp. 1532--1543.

\bibitem{Chen2020MetamorphicTA}
T.~Y. Chen, S.~C. Cheung, and S.-M. Yiu, ``Metamorphic testing: A new approach
  for generating next test cases,'' \emph{ArXiv}, vol. abs/2002.12543, 2020.

\bibitem{Chen2008AnIA}
T.~Y. Chen, J.~W.~K. Ho, H.~Liu, and X.~Xie, ``An innovative approach for
  testing bioinformatics programs using metamorphic testing,'' \emph{BMC
  Bioinformatics}, vol.~10, pp. 24 -- 24, 2008.

\bibitem{Xie2011TestingAV}
X.~Xie, J.~W.~K. Ho, C.~Murphy, G.~E. Kaiser, B.~Xu, and T.~Y. Chen, ``Testing
  and validating machine learning classifiers by metamorphic testing,''
  \emph{The Journal of systems and software}, 2011.

\bibitem{Dwarakanath2018IdentifyingIB}
A.~Dwarakanath, M.~Ahuja, S.~Sikand, R.~M. Rao, R.~P. J.~C. Bose, N.~Dubash,
  and S.~Podder, ``Identifying implementation bugs in machine learning based
  image classifiers using metamorphic testing,'' \emph{Proceedings of the 27th
  ACM SIGSOFT International Symposium on Software Testing and Analysis}, 2018.

\bibitem{Zhang2018DeepRoadGM}
M.~Zhang, Y.~Zhang, L.~Zhang, C.~Liu, and S.~Khurshid, ``Deeproad: Gan-based
  metamorphic testing and input validation framework for autonomous driving
  systems,'' \emph{2018 33rd IEEE/ACM International Conference on Automated
  Software Engineering (ASE)}, 2018.

\bibitem{hateoffensive}
T.~Davidson, D.~Warmsley, M.~Macy, and I.~Weber, ``Automated hate speech
  detection and the problem of offensive language,'' in \emph{Proceedings of
  the 11th International AAAI Conference on Web and Social Media}, ser. ICWSM
  '17, 2017, pp. 512--515.

\bibitem{Song2021EvidenceAN}
K.~Song, Y.~Kang, W.~Gao, Z.~Gao, C.~Sun, and X.~Liu, ``Evidence aware neural
  pornographic text identification for child protection,'' in \emph{AAAI},
  2021.

\bibitem{Mathew2021HateXplainAB}
B.~Mathew, P.~Saha, S.~M. Yimam, C.~Biemann, P.~Goyal, and A.~Mukherjee,
  ``Hatexplain: A benchmark dataset for explainable hate speech detection,'' in
  \emph{AAAI}, 2021.

\bibitem{Kirk2021HatemojiAT}
H.~R. Kirk, B.~Vidgen, P.~R{\"o}ttger, T.~Thrush, and S.~A. Hale, ``Hatemoji: A
  test suite and adversarially-generated dataset for benchmarking and detecting
  emoji-based hate,'' \emph{ACL}, vol. abs/2108.05921, 2021.

\bibitem{Morris2020ReevaluatingAE}
J.~X. Morris, E.~Lifland, J.~Lanchantin, Y.~Ji, and Y.~Qi, ``Reevaluating
  adversarial examples in natural language,'' \emph{EMNLP}, 2020.

\bibitem{Zang2019WordlevelTA}
Y.~Zang, C.~Yang, F.~Qi, Z.~Liu, M.~Zhang, Q.~Liu, and M.~Sun, ``Word-level
  textual adversarial attacking as combinatorial optimization,'' in
  \emph{Annual Meeting of the Association for Computational Linguistics}, 2019.

\bibitem{Zhang2014SearchbasedIO}
J.~M. Zhang, J.~Chen, D.~Hao, Y.~Xiong, B.~Xie, L.~Zhang, and H.~Mei,
  ``Search-based inference of polynomial metamorphic relations,''
  \emph{Proceedings of the 29th ACM/IEEE International Conference on Automated
  Software Engineering}, 2014.

\bibitem{Zhang2019AutomaticDA}
B.~Zhang, H.~Zhang, J.~Chen, D.~Hao, and P.~Moscato, ``Automatic discovery and
  cleansing of numerical metamorphic relations,'' \emph{2019 IEEE International
  Conference on Software Maintenance and Evolution (ICSME)}, pp. 235--245,
  2019.

\bibitem{Carzaniga2014CrosscheckingOF}
A.~Carzaniga, A.~Goffi, A.~Gorla, A.~Mattavelli, and M.~Pezz{\`e},
  ``Cross-checking oracles from intrinsic software redundancy,''
  \emph{Proceedings of the 36th International Conference on Software
  Engineering}, 2014.

\bibitem{notrobustself-driving}
C.~Ziegler, ``A google self-driving car caused a crash for the first time.
  [online],''
  \url{https://www.theverge.com/2016/2/29/11134344/google-self-driving-car-crash-report},
  2016, accessed: 2016-09.

\bibitem{notrobusttesla}
S.~Levin, ``Tesla fatal crash: 'autopilot' mode sped up car before driver
  killed, report finds [online],''
  \url{https://www.theguardian.com/technology/2018/jun/07/tesla-fatal-crash-silicon-valley-autopilot-mode-report},
  2018, accessed: 2018-06.

\bibitem{Carlini2016HiddenVC}
N.~Carlini, P.~Mishra, T.~Vaidya, Y.~Zhang, M.~E. Sherr, C.~Shields, D.~A.
  Wagner, and W.~Zhou, ``Hidden voice commands,'' in \emph{USENIX Security
  Symposium}, 2016.

\bibitem{Tu2021ExploringAR}
J.~Tu, H.~Li, X.~Yan, M.~Ren, Y.~Chen, M.~Liang, E.~Bitar, E.~Yumer, and
  R.~Urtasun, ``Exploring adversarial robustness of multi-sensor perception
  systems in self driving,'' \emph{ArXiv}, vol. abs/2101.06784, 2021.

\bibitem{Luo2021InteractivePF}
Y.~Luo, M.~Meghjani, Q.~H. Ho, D.~Hsu, and D.~Rus, ``Interactive planning for
  autonomous urban driving in adversarial scenarios,'' \emph{2021 IEEE
  International Conference on Robotics and Automation (ICRA)}, pp. 5261--5267,
  2021.

\bibitem{Pei2017DeepXploreAW}
K.~Pei, Y.~Cao, J.~Yang, and S.~S. Jana, ``Deepxplore: Automated whitebox
  testing of deep learning systems,'' \emph{Proceedings of the 26th Symposium
  on Operating Systems Principles}, 2017.

\bibitem{Zhang2022MachineLT}
J.~Zhang, M.~Harman, L.~Ma, and Y.~Liu, ``Machine learning testing: Survey,
  landscapes and horizons,'' \emph{IEEE Transactions on Software Engineering},
  vol.~48, pp. 1--36, 2022.

\bibitem{Riccio2020TestingML}
V.~Riccio, G.~Jahangirova, A.~Stocco, N.~Humbatova, M.~Weiss, and P.~Tonella,
  ``Testing machine learning based systems: a systematic mapping,''
  \emph{Empir. Softw. Eng.}, vol.~25, pp. 5193--5254, 2020.

\bibitem{Humbatova2021DeepCrimeMT}
N.~Humbatova, G.~Jahangirova, and P.~Tonella, ``Deepcrime: mutation testing of
  deep learning systems based on real faults,'' \emph{Proceedings of the 30th
  ACM SIGSOFT International Symposium on Software Testing and Analysis}, 2021.

\bibitem{Pham2021DEVIATEAD}
H.~V. Pham, M.~Kim, L.~Tan, Y.~Yu, and N.~Nagappan, ``Deviate: A deep learning
  variance testing framework,'' \emph{2021 36th IEEE/ACM International
  Conference on Automated Software Engineering (ASE)}, pp. 1286--1290, 2021.

\bibitem{Wang2021RobOTRT}
J.~Wang, J.~Chen, Y.~Sun, X.~Ma, D.~Wang, J.~Sun, and P.~Cheng, ``Robot:
  Robustness-oriented testing for deep learning systems,'' \emph{2021 IEEE/ACM
  43rd International Conference on Software Engineering (ICSE)}, pp. 300--311,
  2021.

\bibitem{Huang2021CoverageGuidedTF}
W.~Huang, Y.~Sun, X.-E. Zhao, J.~Sharp, W.~Ruan, J.~Meng, and X.~Huang,
  ``Coverage-guided testing for recurrent neural networks,'' \emph{IEEE
  Transactions on Reliability}, 2021.

\bibitem{Zhang2022ImprovingAT}
J.~Zhang, W.~Wu, J.~tse Huang, Y.~Huang, W.~Wang, Y.~Su, and M.~R. Lyu,
  ``Improving adversarial transferability via neuron attribution-based
  attacks,'' \emph{2022 IEEE/CVF Conference on Computer Vision and Pattern
  Recognition (CVPR)}, pp. 14\,973--14\,982, 2022.

\bibitem{Madry2018TowardsDL}
A.~Madry, A.~Makelov, L.~Schmidt, D.~Tsipras, and A.~Vladu, ``Towards deep
  learning models resistant to adversarial attacks,'' \emph{ICLR}, vol.
  abs/1706.06083, 2018.

\bibitem{Asyrofi2021CanDT}
M.~H. Asyrofi, Z.~Yang, J.~Shi, C.~W. Quan, and D.~Lo, ``Can differential
  testing improve automatic speech recognition systems?'' \emph{2021 IEEE
  International Conference on Software Maintenance and Evolution (ICSME)}, pp.
  674--678, 2021.

\bibitem{Gao2020FuzzTB}
X.~Gao, R.~K. Saha, M.~R. Prasad, and A.~Roychoudhury, ``Fuzz testing based
  data augmentation to improve robustness of deep neural networks,'' \emph{2020
  IEEE/ACM 42nd International Conference on Software Engineering (ICSE)}, pp.
  1147--1158, 2020.

\bibitem{Ma2018MODEAN}
S.~Ma, Y.~Liu, W.-C. Lee, X.~Zhang, and A.~Y. Grama, ``Mode: automated neural
  network model debugging via state differential analysis and input
  selection,'' \emph{Proceedings of the 2018 26th ACM Joint Meeting on European
  Software Engineering Conference and Symposium on the Foundations of Software
  Engineering}, 2018.

\bibitem{Tao2020TRADERTD}
G.~Tao, S.~Ma, Y.~Liu, Q.~Xu, and X.~Zhang, ``Trader: Trace divergence analysis
  and embedding regulation for debugging recurrent neural networks,''
  \emph{2020 IEEE/ACM 42nd International Conference on Software Engineering
  (ICSE)}, pp. 986--998, 2020.

\bibitem{Zhang2017SentimentAA}
L.~Zhang and B.~Liu, ``Sentiment analysis and opinion mining,'' in
  \emph{Encyclopedia of Machine Learning and Data Mining}, 2017.

\bibitem{Wang2017EmotionRW}
S.~Wang, W.~Wang, J.~Zhao, S.~Chen, Q.~Jin, S.~Zhang, and Y.~Qin, ``Emotion
  recognition with multimodal features and temporal models,'' \emph{Proceedings
  of the 19th ACM International Conference on Multimodal Interaction}, 2017.

\bibitem{Bahdanau2015NeuralMT}
D.~Bahdanau, K.~Cho, and Y.~Bengio, ``Neural machine translation by jointly
  learning to align and translate,'' \emph{ICLR}, vol. abs/1409.0473, 2015.

\bibitem{Wang2022UnderstandingAI}
W.~Wang, W.~Jiao, Y.~Hao, X.~Wang, S.~Shi, Z.~Tu, and M.~R. Lyu,
  ``Understanding and improving sequence-to-sequence pretraining for neural
  machine translation,'' in \emph{Annual Meeting of the Association for
  Computational Linguistics}, 2022.

\bibitem{Jiao2022TencentsMM}
W.~Jiao, Z.~Tu, J.~Li, W.~Wang, J.~tse Huang, and S.~Shi, ``Tencent’s
  multilingual machine translation system for wmt22 large-scale african
  languages,'' \emph{WMT}, 2022.

\bibitem{Wang2017TacotronTE}
Y.~Wang, R.~J. Skerry-Ryan, D.~Stanton, Y.~Wu, R.~J. Weiss, N.~Jaitly, Z.~Yang,
  Y.~Xiao, Z.~Chen, S.~Bengio, Q.~V. Le, Y.~Agiomyrgiannakis, R.~A.~J. Clark,
  and R.~A. Saurous, ``Tacotron: Towards end-to-end speech synthesis,'' in
  \emph{Interspeech}, 2017.

\bibitem{Ma2018FPETSFP}
D.~Ma, Z.~Su, W.~Wang, and Y.~Lu, ``Fpets: Fully parallel end-to-end
  text-to-speech system,'' in \emph{AAAI Conference on Artificial
  Intelligence}, 2018.

\bibitem{Gupta2020MachineTT}
S.~Gupta, ``Machine translation testing via pathological invariance,''
  \emph{2020 IEEE/ACM 42nd International Conference on Software Engineering:
  Companion Proceedings (ICSE-Companion)}, pp. 107--109, 2020.

\bibitem{He2021TestingMT}
P.~He, C.~Meister, and Z.~Su, ``Testing machine translation via referential
  transparency,'' \emph{2021 IEEE/ACM 43rd International Conference on Software
  Engineering (ICSE)}, pp. 410--422, 2021.

\bibitem{Jiao2023IsCA}
W.~Jiao, W.~Wang, J.~tse Huang, X.~Wang, and Z.~Tu, ``Is chatgpt a good
  translator? a preliminary study,'' \emph{ArXiv}, vol. abs/2301.08745, 2023.

\bibitem{Ribeiro2020BeyondAB}
M.~T. Ribeiro, T.~S. Wu, C.~Guestrin, and S.~Singh, ``Beyond accuracy:
  Behavioral testing of nlp models with checklist,'' in \emph{ACL}, 2020.

\bibitem{Ahlgren2021TestingWE}
J.~Ahlgren, M.~E. Berezin, K.~Bojarczuk, E.~Dulskyte, I.~Dvortsova, J.~George,
  N.~Gucevska, M.~Harman, M.~Lomeli, E.~Meijer, S.~Sapora, and
  J.~Spahr-Summers, ``Testing web enabled simulation at scale using metamorphic
  testing,'' \emph{2021 IEEE/ACM 43rd International Conference on Software
  Engineering: Software Engineering in Practice (ICSE-SEIP)}, pp. 140--149,
  2021.

\bibitem{Kapoor2019MindYL}
R.~Kapoor, Y.~K. Singla, K.~Rajput, R.~R. Shah, P.~Kumaraguru, and
  R.~Zimmermann, ``Mind your language: Abuse and offense detection for
  code-switched languages,'' \emph{AAAI}, 2019.

\bibitem{Cid2008TheIO}
I.~Cid, L.~R. Janeiro, J.~R. M{\'e}ndez, D.~Glez-Pe{\~n}a, and F.~F. Riverola,
  ``The impact of noise in spam filtering: A case study,'' in \emph{ICDM},
  2008.

\bibitem{Li2021EnhancingMR}
J.~Li, T.~Du, X.~Liu, R.~Zhang, H.~Xue, and S.~Ji, ``Enhancing model robustness
  by incorporating adversarial knowledge into semantic representation,''
  \emph{ICASSP 2021 - 2021 IEEE International Conference on Acoustics, Speech
  and Signal Processing (ICASSP)}, pp. 7708--7712, 2021.

\end{thebibliography}
\end{document}